# A General Framework for Saliency Detection Methods

Fateme Mostafaie[1], Zahra Nabizadeh[1], Nader Karimi[1], Shadrokh Samavi[2]
[1]Department of Electrical and Computer Engineering, Isfahan University of Technology, Isfahan, Iran
[2]Department of Electrical and Computer Engineering, McMaster University, Canada

*Abstract*—**Saliency detection is one of the most challenging problems in image analysis and computer vision. Many approaches propose different architectures based on the psychological and biological properties of the human visual attention system. However, there is still no abstract framework that summarizes the existing methods. In this paper, we offered a general framework for saliency models, which consists of five main steps: pre-processing, feature extraction, saliency map generation, saliency map combination, and post-processing. Also, we study different saliency models containing each level and compare their performance. This framework helps researchers to have a comprehensive view of studying new methods.**

## I. INTRODUCTION

Saliency detection aims to determine the most salient and informative regions in an image that attract the human's eyes. Many saliency detection algorithms have recently been generated to simulate visual attention mechanisms. The human visual system has top-down (task-driven) and bottom-up (stimulus-driven) mechanisms [1]. Saliency detection methods are classified into two main categories: unsupervised and supervised approaches. The unsupervised methods are proposed according to the biological and psychological attributes of the human visual system. The supervised methods use machine learning technology to implement the saliency detection model [2]. Early models focus on unsupervised methods and aim to find objects with different visual features from their surrounding area. They use simple features like color, edge, and texture rather than complex ones like shapes and objects. As machine learning improved, supervised methods became more popular because they performed better than unsupervised ones [3].

Recently, by creating numerous datasets and the remarkable success of deep convolutional neural networks (CNNs), many deep models have been proposed with better performance than traditional methods [4]. Nowadays, saliency detection is being used as a pre-processing step in many computer vision and image processing tasks, such as image segmentation [5], object recognition [6], video/image foreground co-segmentation [7], and image retargeting [8].

Early research for saliency detection focused on predicting human eye fixation. In [9], three feature maps are extracted based on the input image's color, intensity, and orientation and are combined to generate a final saliency map. In [10], the model of [9] is extended, and a method based on a Markov chain on the fully-connected map and a graph-based dissimilarity measure are proposed. Work in [11] proposed a saliency detection model, which estimates image saliency by learning the prior knowledge gained from millions of images.

In computer vision, it is offered that human attention got attracted by salient objects more than some isolated fixations [12]. Work in [13] proposed a model which uses a set of features, including multi-scale contrast, center-surround histogram, and spatial color distribution to describe a salient object locally, regionally, and globally. Work in [14] utilized random walks on a graph modeled by color and orientation features to extract salient regions. In [15], Perazzi et al. presented a model based on image abstraction into representative elements and contrast-based saliency measures to produce a per-pixel saliency map.

As introduced before, many saliency detection methods are suitable for various applications. However, no general framework provides a global view of a saliency detection model's structure. This paper aims to cover recent saliency detection approaches and propose a general and abstract framework developed based on their structure. It has been tried to summarize primary and new saliency detection approaches. Therefore, the details of some methods may not be mentioned. The proposed framework contains five steps of the saliency detection algorithm, including pre-processing, feature extraction, saliency map generation, and map combination. These steps are shared among the reviewed approaches. In the following, each step will be explained clearly in Section II; Section III describes the conclusion.

## II. PROPOSED METHOD

It may not be possible to design a framework containing all saliency detection models. However, most of the saliency models can be summarized by five blocks shown in Fig. 1. The most vital block is feature extraction, which image can be fed into it directly or after enhancement by a pre-processing block. In the feature extraction phase, various features are extracted according to the method of implementing the intended saliency detection model. Then these features are passed into a saliency map generation block, which produces several saliency maps from a different group of features. In the next step, the map combination block tries to fuse different saliency maps, resulting in the former block producing a final map. In the end, the post-processing block can improve the final saliency map. It is crucial to note that some saliency detection models may not have some of these blocks.

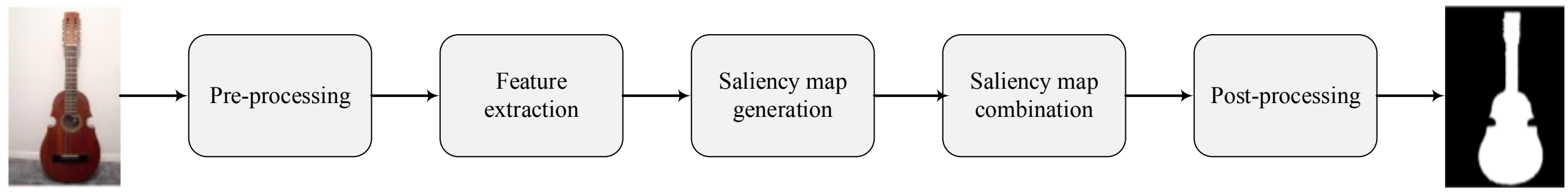

Fig. 1. General framework for saliency detection

### A. pre-processing

Sometimes, input images need to be changed to be more suitable to be fed into the saliency detection models. Works in [16],[17],[18] decompose the input image into non-overlapping regions to extract features from each image region separately. In [2], resizing the input image is done to decrease the resolution of the image. Consequently, it helps to reduce training and testing time considerably. Work in [19] generates pseudo-colored MRI for detecting tumor regions in multi-sequence MR images because the pseudo-coloring method makes the input image proper for future processing. However, in some input images, distracting areas have high diversity compared to their surroundings. Hence, they can attract the attention of the saliency models [20]. In [20], Xiao et al. developed a distraction detection network to remove distracting regions from the input image. Therefore, a better saliency prediction could derive from distraction-free input.

### B. Feature extraction

Two types of features are extracted from the image: local and global. Global elements help model the whole image's saliency, and local features help to find the saliency of areas with more features [3]. Each feature can be extracted from the input or processed image depending on the strategies used to solve the saliency detection problem. Besides, some methods used complement features extracted from different network layers. As shown in Fig. 2, generally, we categorize extracted features in saliency models in three blocks: local features, global features, and a combination of local and global features. In the following, we describe them in more detail.

*1) Local features:* Local features describe the image patches of an object. Region-based feature extraction is one method for extracting local features. Saliency detection should not be pixel-wise but based on looking at an isolated pixel. Instead, it can use region-based feature extraction to compute the correct saliency score. Over-segmenting the input image into non-overlapping regions could allow us to calculate specific information about each area. Some regional features are as follows: Average normalized coordinates, bounding box location, variances of the RGB values, and the aspect ratio of the bounding box. Also, in [16], the relationship between a regional feature based on semantic priors and their saliency value is studied. Work in [17] benefits from three deep CNNs to capture multi-scale features from each region. For every part, the features are extracted from three nested and increasingly larger rectangle windows, the bounding box of the considered area, the bounding box of its immediate neighboring regions, and the entire image. There is the same approach in [18], which extracted each region's feature vectors from three nested windows to discover segment-level visual contrast. Saliency is a unique property in a specific object, which makes it different from its surrounding area [3].

In some saliency detection models, different local features for different aims are extracted. It was suggested in [3] to add contrast features associated with each local feature to capture the difference between each feature and its surrounding region. Work in [2] adopted a sparse feature extraction method to discover a high contrast pattern. First, the input image is turned into a multi-scale patch-based representation. For each, the Independent Component Analysis (ICA) is applied to generate a complete sparse dictionary of weights. The feature vector can compute based on these weights.

*2) Global features:* Global features describe the image as a whole to generalize the entire object. Semantic information could help this operation notably. Although saliency detection models have improved significantly, there is still a considerable gap between model prediction and what a human perceives from an image. This gap is due to these models' lack of semantic content information [21]. Researches about the human visual system show that people get attracted more to the main objects of the pictures than other regions. As a result, the efficiency of models depends on their ability to discover semantic information [2].

Additionally, images in the recent dataset have multiple objects with a complex background; hence, the models should use semantic features to detect which objects are salient [16]. In [15], a fully convolutional network (FCN) is used to extract semantic maps related to different semantic classes. Work in [2] utilized the feature maps from convolutional layers of VGG; therefore, it leads to describing images at different scales and views. Extracted feature maps from deeper layers convey information about objects, while obtained maps from shallower ones convey structural details such as edges and textures. Accordingly, extracted feature maps from deeper layers help to detect the semantic objects in input images. In addition to semantic information, positional information is helpful for general objects. Work in [23] used this benefit to detect diabetic macular edema (DME) in the retinal image because DME always appears in a particular region. The statistical analyzing method is used to get the position property from the ground truth.

*3) Combination of Local and Global Features:* As mentioned before, there are two categories of saliency methods – bottom-up and top-down. Bottom-up models compute saliency maps based on low-level features like color, intensity, and texture. At the same time, top-down approaches adopt high-level information such as object and face detectors [17]. Unlike the shallow layers of the network, deep layers capture high-level semantic information but messy data. By combining features of shallow and deep layers, we can meet this requirement, but it may cause reducing the confidence in the prediction of deep layers [22]. Zhiming [3] implemented a grid-like Convolutional Neural Network (CNN) to extract multi-scale local feature maps from its intermediate convolutional layers to reach this goal. Also, [22] tries to capture residual features, like object boundaries and other undetected object parts. It adopted VGG-16 and select {conv1-2, conv2-2, conv3-3, conv4-3, and conv5-3} as side-outputs. Then residual features are extracted from each side output. Work in [24] benefits from an encoder network to extract a rich feature representation from the input image and refinement stages, which recover lost contextual information. In the refinement stages, the spatial resolution that is lost at the deepest layer of the encoder gradually recovers from earlier representations, so an accurate prediction is achieved. In [20], short- and long-range connections are introduced to combine features at the same scale and transport features to other scales. They help to gather both global and local features extracted from different layers of networks.

### C. Saliency map generation

When the feature extraction process is finished, it is time to obtain saliency maps from feature maps. Generally, we categorize saliency detection models into two groups: those that produce several saliency maps in different ways and the others that generate just one saliency map (see Fig. 3). In the following, they will be explained more with some examples.

*1) Models with multipath saliency map generation:* In [16], Tam V Nguyen proposed explicit and implicit saliency maps using semantic priors. An explicit saliency map aims to detect the semantic class of an image that is more noticeable for people at first glance, and some mathematical computation generates it on semantic maps. Despite this, an implicit saliency map is used to detect objects, not in semantic classes. It is produced by training a regressor, which maps the extracted features to the saliency values. Work in [2] proposed a saliency framework that consists of two pathways, i.e., semantic-aware saliency and contrast-aware saliency. The target of a semantic-aware saliency path is to capture semantic information about the image, like objects or object parts. In contrast, the contrast-aware saliency path tries to identify feature contrast. The contrast-aware saliency map is computed based on multi-scale adaptive sparse representation and information maximization. However, a semantic-aware saliency map is generated by a two-step procedure based on the VGG network.

Detecting the salient object in the image requires first finding the image's global context and assigning saliency value to the small region later [3]. To fulfill this aim, work in [3] introduced local and global saliency maps. The global saliency map is derived from the last convolutional layer of the VGG-16 network. However, the local saliency map is extracted from all its convolutional layers. Up-sampling is done to prevent producing a coarse map, step by step. Work in [18] proposed a pixel-level fully convolutional stream and a segment-level spatial pooling stream. The segment-level spatial pooling stream aims to detect visual contrast between a region and its neighborhoods. Nevertheless, pixel-level fully convolutional stream captures visual contrast and semantic features. It uses VGG-16 as a pre-trained network to design an end-to-end convolutional network, directly making a pixel-level saliency map from the input image.

*2) Models with mono saliency map generation:* In [17], three feature vectors are extracted for each region, then concatenated and fed into a neural network. This network estimates the saliency value of each image region. To use extracted features from both deep and shallow layers, work in [22] extracted five feature maps from five convolutional layers of VGG-16. Then the saliency map predicted in the deeper layer was refined by residual features step by step. Also,[20] proposed a model in which encoded features are up-sampled with low-level features via dense connections, so multi-scale features are extracted, and the accuracy of the saliency map is increased.

### D. Saliency map combination

As it is mentioned, some models extract different saliency maps from different streams. Hence they should be merged to create the final saliency map. Fusing different saliency maps, which are driven from different ways, are a challenging task. In [2], two different saliency maps are integrated by summing them up after maxima normalization. In [16],[3], a linear combination of obtained saliency maps makes the final saliency map. Also, work in [18], [24] used a convolutional layer to fuse generated saliency maps.

### E. Post-processing

After constructing the final saliency map, some methods could be used to enhance it. To increase the performance of the saliency detection model, in [1], Jianjun Lei et al. apply an iterative optimization process to gain an optimal saliency map from the existing map. Moreover, expanding the salient region is used in some methods. When we look at an image, we focus not only on the main object but also on some areas around it that may include important information. As a result of this concept, the region that contains the salient object could be recognized at first and then extended to its surrounding context [25]. By weighting each pixel's saliency, based on Euclidean distance, the information has the most substantial influence on the nearest pixel, and the salient region is extended [25]. Work in [23] utilizes a Gaussian filter to obtain an enhanced saliency map.

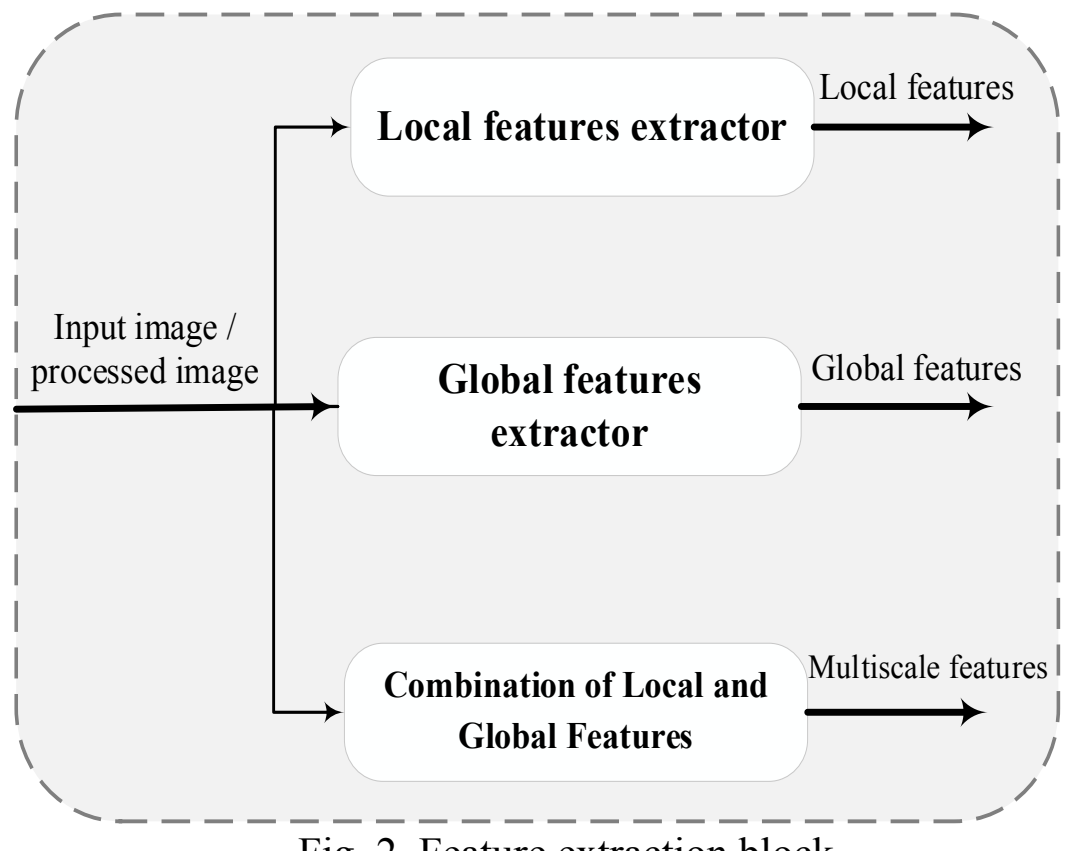

Fig. 2. Feature extraction block

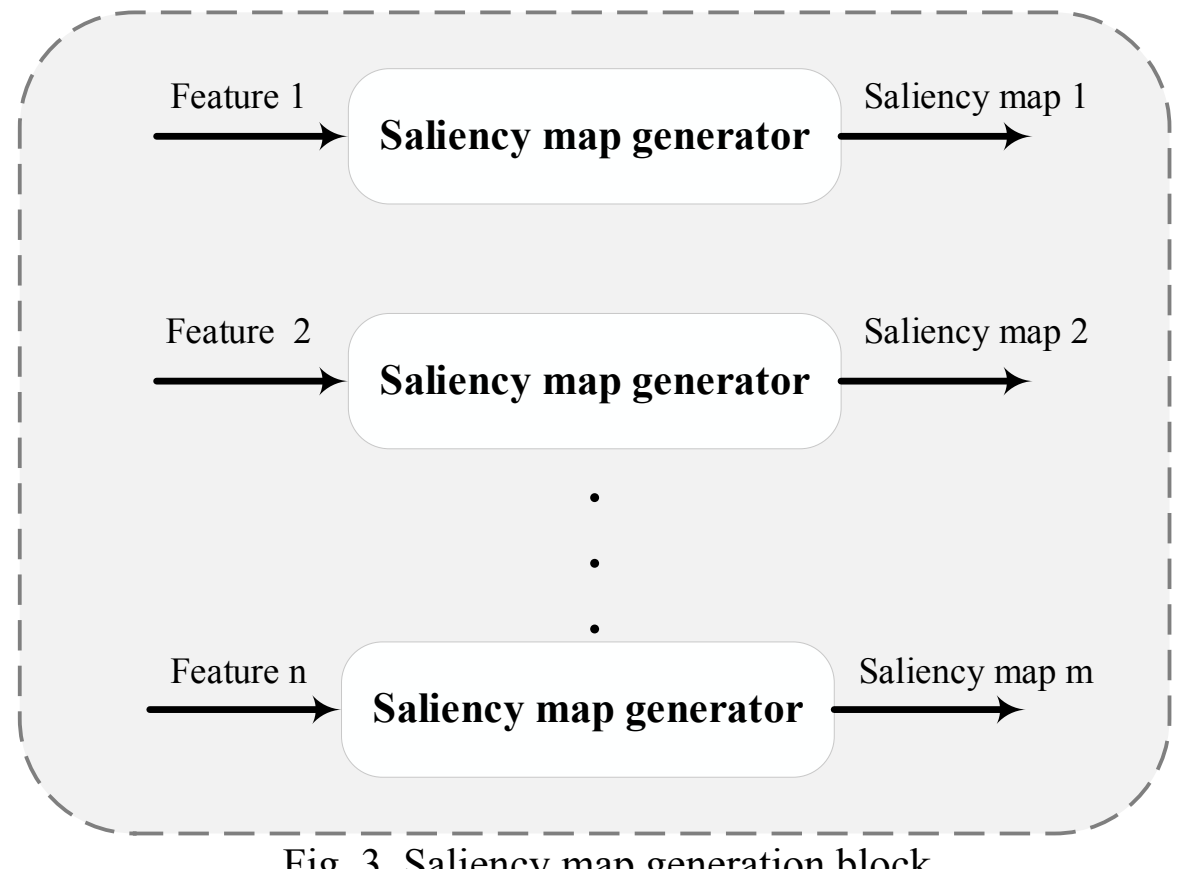

Fig. 3. Saliency map generation block

## III. CONCLUSION

In this paper, we proposed not only an abstract but also a complete framework for saliency detection models, which helps researchers to achieve a professional view to realize the fundamental steps of a saliency detection model. Our general saliency detection framework consists of five main blocks, including pre-processing, feature extraction, saliency map generation, saliency map combination, and post-processing. We further explain each block in a separate part and study various methods which contain that block.

The first block is pre-processing, which prepares the image and makes it more proper for the feature extraction phase. The second is the extraction block, which extracts features from an image to generate a saliency map. These features are chosen based on the saliency detection approach. The next block is the saliency map generation unit, which uses extracted features from the previous block to compute the saliency map. Some methods may produce more than one saliency map in this step to achieve high accuracy. In other words, they use different saliency maps derived from various features to benefit from complement features. The Fourth block is a saliency map combination that fuses different saliency maps generated in the previous block. The final block is post-processing, which enhances the generated saliency map and makes it more accurate.

To understand the role of each block in the performance of studied methods, we made Table I to compare their performance. Three types of datasets are used in papers in this table, (i) PASCAL-S, (ii) DUT-OMRON, and (iii) HKU-IS. The PASCAL-S dataset has 850 images from PASCAL 2010 in 12 subjects. This dataset includes original images, full segmentation, eye fixations, and salient object masks. The DUT-OMRON dataset consists of 5,168 high-quality images with one or more salient objects and a relatively complex background. It includes original images, pixel-wise ground truth, bounding box ground truth, and eye-fixation ground truth of it. The HKU-IS dataset is [17] 's new dataset which contains the original image with pixel-wise annotation of salient objects. We use two metrics for comparing all methods, (i) the F-measure and (ii) the mean absolute error (MAE). The reported F-measure score for each method is shown in red. Larger F scores are better, while the reported values of MAE scores are shown in blue, where smaller numbers are better.

TABLE I: Comparing Different Methods

| Paper | Pre-processing | Feature extraction | Saliency map generation (network) | Saliency map combination | Post-processing | Dataset | Score (F-measure MAE) |
|---|---|---|---|---|---|---|---|
| [20] | Removing distracting regions | Combination of local and global features | D-Net: VGG-16 S-Net: Network with dense short- / long-range connections | * | * | PASCAL-S | 0.845 0.103 |
| | | | | | | DUT-OMRON | 0.770 0.118 |
| [22] | * | Combination of local and global features | VGG-16 | * | * | PASCAL-S | 0.818 0.106 |
| | | | | | | DUT-OMRON | 0.762 0.071 |
| [24] | * | Combination of local and global features | Resnet-101 | Convolutional layer | * | PASCAL-S | 0.873 0.091 |
| [18] | Over-segmenting | Global features/ Local features | Pixel-Level: VGG-16 Segment-Level: NN-Layer | Convolutional layer | * | PASCAL-S | 0.815 0.113 |
| | | | | | | DUT-OMRON | 0.733 0.084 |
| [3] | * | Combination of local and global features | Grid- CNN Network | Linear combination | * | PASCAL-S | 0.831 0.099 |
| | | | | | | DUT-OMRON | 0.753 0.080 |
| [16] | Over-segmenting | Global features/ Local features | random forest regressor | Linear combination | * | HKU-IS | 0.86(approx.) 0.1(approx.) |
| [17] | Over-segmenting | Local features | CNN | * | * | HKU-IS | 0.81(approx.) 0.080(approx.) |

The best score in PASCAL-S dataset for both metrics is related to [24], which used extracted features from different layers to refine the saliency map. The best F-measure scores in the DUT-OMRON dataset are related to [20] and [22]. It shows that gathering high-level and low-level features is an effective technique for complex images. In [20], multi-scale features are extracted by adopting dense connections in the saliency detection network, and [22] proposed a reverse attention block that used the prediction of a deeper layer to gradually improve the resolution of the saliency map. In the HKU-IS dataset, [16] has a better F-measure score using the image's semantic content, while [17] gets a better MAE score by utilizing multi-scale features from each region.

In general, it could be concluded that extracting features from different scales of the image provides a global and local view, and it results in generating a better saliency map.